\documentclass[runningheads]{llncs}
\usepackage[T1]{fontenc}
\usepackage{graphicx}

\usepackage{hyperref}
\begin{document}
\title{Explainability of Deep Learning-Based Plant Disease Classifiers Through Automated Concept Identification}
\titlerunning{Explainability of DL-based Plant Disease Classifiers}
%
%
%
%
\author{Jihen Amara\inst{1,2}\orcidID{0000-0003-3675-259X}  \and
Birgitta König-Ries\inst{1,2}\orcidID{0000-0002-2382-9722}\and
Sheeba Samuel\inst{3}\orcidID{0000-0002-7981-8504}}
\authorrunning{J. Amara et al.}
%
\institute{Heinz Nixdorf Chair for Distributed Information Systems, Friedrich Schiller University Jena, Jena, Germany \and
Michael Stifel Center Jena, Jena,  Germany  \and
Distributed and Self-organizing Systems, Chemnitz University of Technology, Chemnitz, Germany\\
\email{(jihene.amara, birgitta.koenig-ries)@uni-jena.de\\
sheeba.samuel@informatik.tu-chemnitz.de } \\
}

\maketitle              
\begin{abstract}
While deep learning has significantly advanced automatic plant disease detection through image-based classification, improving model explainability remains crucial for reliable disease detection. 
In this study, we apply the Automated Concept-based Explanation (ACE) method to plant disease classification using the widely adopted InceptionV3 model and the PlantVillage dataset. ACE automatically identifies the visual concepts found in the image data and provides insights about the critical features influencing the model predictions. This approach reveals both effective disease-related patterns and incidental biases, such as those from background or lighting that can compromise model robustness.  Through systematic experiments, ACE helped us to identify relevant features and pinpoint areas for targeted model improvement.  Our findings demonstrate the potential of ACE to improve the explainability of plant disease classification based on deep learning, which is essential for producing transparent tools for plant disease management in agriculture.

\end{abstract}
\section{Introduction}

Agriculture is primordial for human life. It is the foundation for food production and economic stability around the world. With the global population expected to reach 10 billion by 2050, food production must increase by 60\% to meet growing demand \cite{ristaino2021persistent}. To achieve this, it is required to boost crop yield quality and minimise food loss caused by diseases. Hence, efficient and innovative agricultural practices are more important than ever.

Consequently, in recent years, we have noticed the emergence of automated plant disease detection methods that can save time and provide better help and support to farmers and other stakeholders.

Central to these innovations is deep learning (DL), a powerful branch of AI that enables the rapid and accurate identification of plant diseases through image analysis. By detecting diseases early and with greater precision, deep learning enhances the effectiveness of these  automated systems. This can allow for more timely and targeted interventions, ultimately helping to save crops and reduce losses. 
Different works have applied deep learning for plant disease classification and demonstrated its efficiency. They often employ convolutional neural networks (CNN) such as inceptionV3 \cite{szegedy2015going,fan2022leaf,lee2020new,qiang2019identification,subramanian2022fine}, which are suitable for image classification tasks.

Even though deep learning methods have advanced image-based plant disease classification, they often lack explainability, which makes users and involved stakeholders question their results.
Deep learning models are usually referred to as a black box due to the complexity of their architectures and the lack of insight into how they reach their decisions. This opacity is a huge problem in critical
domains like healthcare, finance, or agriculture, where decisions come with high stakes and the threat of significant financial losses.

Hence, understanding and explaining the rationale behind the model predictions is essential for farmers and agricultural experts  to trust the model's output especially when it is related to disease management strategies. 
Also, explainability can help in detecting and preventing unexpected biases, especially when dealing with noisy and unbalanced datasets which is often the case in plant disease classification \cite{barbedo2022deep}.

One of the most popular explainability methods is the visual-based approach which helps users understand model decisions by visually highlighting the areas or features in the input data that contribute most to the model's predictions (see Section \ref{sec:related}).
Even though these visual based explanation methods provide useful visual feedback, they often fail to capture more abstract or conceptual information that influences model decision-making.

To address this limitation, we propose the use of concept-based explainability methods.
Specifically, Automated Concept-based Explanation (ACE) \cite{ghorbani2019towards} stands out as a method that can identify and group high-level concepts directly from the data. 
Unlike visual based methods, ACE can help by quantifying the importance of high-level human-interpretable concepts such as leaf color, shape, or disease shape and texture in the model's decision-making process.

This helps agriculture experts check whether the model is generating outcomes based on valid biologically relevant features.
Also, we argue that ACE could not only be a great tool to explain the model decision but can be a valuable method to explore biases and detect irrelevant concepts in the dataset. In some cases, the model may start learning irrelevant patterns or correlations to make its decision due to biases in data collection (i.e background noise). This can result in poor generalisation and decreased accuracy in real-world applications.

Finally, our contributions can be summarised as follows:
\begin{itemize}
  \item We provide new and valuable insights into the DL-based plant disease classification models by adapting the ACE method to automatically identify key visual patterns and concepts used in the model's decision-making process. This approach allows plant and agricultural experts to better understand the model's behaviour, raising greater trust in its predictions.
  
  \item We introduce a new approach that reveals biases in DL model-based plant disease classification and identifies potential issues within the dataset using ACE.  This approach helps diagnose if the model's decisions rely on unwanted correlations instead of meaningful plant disease symptoms. This helps to address and mitigate these issues in future models and datasets toward a more robust and generalizable plant disease classification.
  
  \item We offer technical insights into the application of ACE in the context of plant disease classification, providing a crucial tool for identifying and addressing unique challenges within this field. To the best of our knowledge, this is the first work to apply ACE in this domain, establishing an important step toward improving the reliability and explainability of DL models for plant disease diagnosis.
\end{itemize}

The remainder of this paper is organised as follows. Section 2 presents related work on explainability and its application to plant disease classification. Section 3 details the model, ACE theory, and dataset used in this study. Section 4 presents our experimental results, including discussions on prediction accuracy and model explanation with ACE. Finally, Section 5 offers concluding remarks and possible future work.

\section{Related work}
\label{sec:related}
Several explainability methods have been proposed, with the most common approaches falling into two categories: visual-based explanations and concept-based explanations. In this section, we will describe each set of methods and provide examples of their use in the context of plant disease classification.

Visual-based explanation \cite{zhou2016learning,selvaraju2020grad,ribeiro2016should} generates visual cues highlighting the parts of an input image that contribute most to the model’s prediction \cite{mohamed2022review}. These methods produce heat maps or saliency maps that indicate which areas of an image the deep model considers important when making its predictions.

Zhou et al. \cite{zhou2016learning} proposed Class activation mapping (CAM), a technique that replaces the fully connected layers at the end of a CNN with a global average pooling layer (GAP), applied on the last convolutional feature maps. The CAM is then computed as a weighted linear sum of these feature maps, where the weights are determined by the output class probabilities of the CNN. This resulted in a heatmap that highlighted the regions of the image that were most strongly associated with the predicted class. Selvaraju et al. \cite{selvaraju2020grad} proposed Gradient-weighted Class Activation Mapping (Grad-CAM), a generalisation of CAM that can work with any type of CNN to produce local explanations. This is in contrast to CAM, which specifically needs global average pooling. LIME was proposed by Ribeiro et al. \cite{ribeiro2016should}. It generates explanations by perturbing different parts of the input image (e.g, removing or altering sections) and observing how these changes affect the model's prediction. This approach produces a visual explanation, typically in the form of a heatmap, that identifies the image regions most influential in the model’s decision-making process.
 
In plant disease classification, this form of explanation is particularly useful because it allows researchers and agricultural experts to visually inspect which features of a diseased plant (e.g., spots and discoloration areas) the model is using to make its diagnosis.
 Kinger et al. \cite{kinger2021explainable} presents a  review of visual explanation methods and their application in plant leaf disease detection. The authors started by defining the key concept of explainability. They then discussed recent advancements, focusing on popular methods such as LIME and GradCAM. Then,  using the PlantVillage dataset, the authors fine-tuned the VGG16 model and applied Grad-CAM, Grad-CAM++, and LIME to evaluate the effectiveness of each technique in explaining the model's predictions. Their findings suggest that these visualisation techniques can help farmers better understand the models predictions and make more informed decisions.
 
Additionally, in \cite{hernandez2024field}, Grad-CAM was used to enhance the explainability of automated grapevine downy mildew disease classification. It helped in visualising the critical image areas that influences the model’s decision-making process such as the symptomatic regions like oil spots on grapevine leaves associated with downy mildew.  The approach allowed for a more transparent understanding of how the model identifies disease indicators and accordingly increases confidence in its practical application for field diagnosis. Also, \cite{mehedi2022plant} explored a transfer learning approach with three pretrained models  EfficientNetV2L, MobileNetV2, and ResNet152V2 to detect plant diseases using the PlantVillage dataset. EfficientNetV2L performs best with 99.63\% accuracy. They hence used LIME to explain the decisions of the model. However, LIME explanations can be misleading as they may focus on features that are not actually important for the model's prediction. Toda et al. \cite{toda2019convolutional} investigated CNNs’ predictions for plant disease classification using various visualisation methods. They found that the CNNs can capture the colour and textures of lesions specific to respective diseases. They also found that some layers were not contributing to the inference and then removed them without affecting the classification accuracy.
In \cite{brahimi2019deep}, an alternative approach addressed the explainability problem through a Teacher-student paradigm, trained jointly using multi task learning. The shared representations between the models were used to visualise key image regions critical for classification.This approach produced  sharper visualisations compared to other existing methods. However, this method was computationally and time intensive. In another study,  \cite{ghosal2018explainable} presented a new explanation approach by identifying the top-k high-resolution feature maps that contribute most to the model’s predictions. They found that the highlighted visual features were closely aligned with those used by experts to assess disease severity. However, this method did not reveal the model’s internal mechanisms in detail.
Even though these visual based explanation methods have been frequently used in plant disease DL-based classification literature, they suffer from some key disadvantages. One example is the lack of specificity where the method often highlights a large region of the images without clearly indicating why they are essential for the model which leads to subjectivity and differences of interpretation between different users.
Another example is their fragility and sensitivity to adversarial perturbations \cite{ghorbani2019interpretation}. Where small changes in input can cause the method to focus on entirely different regions which undermine their reliability. Also, other studies have highlighted the potential unreliability of these methods \cite{kindermans2019reliability}. Since they generate importance maps specific to individual input samples, they provide only local explanations, lacking a comprehensive view of the model’s overall behaviour. Additionally, one of their significant drawbacks is the lack of clarity and expressiveness for users. For example, the influence of a single pixel on classification offers little meaningful insight, and the interpretability becomes more complex with a large number of features \cite{molnar2020interpretable}. This reliance on pixel-level information makes it limiting where higher-level concepts (such as shapes or texture) are more relevant for understanding the model decisions.  To overcome these limitations, concept-based approaches were introduced.
 
Concept based explanation methods \cite{kim2018interpretability,ghorbani2019towards} offer a more abstract interpretation of what the model is using to make its decision in terms of higher-level human understandable concepts.
These concepts could be predefined or learned and range from a simple color to an object or a complex idea \cite{holzinger2023toward}. This is suitable for plant disease classification, where it is important to ensure that the model recognizes disease-specific concepts  (e.g., leaf spots, discoloration, or texture) rather than irrelevant background features.

One of these methods is the TCAV approach, which was proposed by \cite{kim2018interpretability}. It explains how a model makes predictions by examining the influence of user-defined concepts on the model’s decisions. A key component within TCAV are the concept activation vectors (CAVs) that represent specific concepts (such as “stripes” or “colors”) in the model hidden layers. Using these CAVs, TCAV can measure the importance of the defined concept to the model for the prediction globally. More details about this will be given in Section \ref{sec:ace}.

Our previous work \cite{amara2023concept} provided an initial exploration into understanding the semantic concepts that CNNs learn during plant disease diagnosis using TCAV. This study was the first to collect various related disease-based concepts to analyse CNN interpretations in this context.

Even though TCAV has proven its usefulness in defining and testing specific concepts such as discolorations, disease patterns, and symptoms  (i.e., blotchiness,  Crackedness, and wrinkledness) \cite{amara2023concept}, it requires users to have a good understanding of which concepts are the most relevant and to have enough resources to collect adequate examples. 

This manual approach could be a massive burden in plant disease deep learning-based classification, especially when users may not know which concepts are the most relevant for distinguishing between diseases.

Also, they may lack sufficient labelled data to define these concepts effectively. 
This is where Automated Concept-based Explanation (ACE) \cite{ghorbani2019towards} offers a significant advantage. ACE was proposed in \cite{ghorbani2019towards} to tackle the problem above. It works on automatically extracting relevant concepts directly from the target class images without the need for manual interference. More details on ACE are provided in Section \ref{sec:ace}.

This automatization is especially valuable for plant disease diagnosis, where visual symptoms are not always well-defined. 
We argue that ACE will help in revealing new concepts that could be important for the prediction but were unnoticed before. This can reduce the problem of mismatch between the concepts a user defines and the actual concepts the model uses to make a decision.
This conflict becomes evident in situations where the dataset is imbalanced or contains mislabeled instances, which is often the case in real-world plant disease datasets \cite{barbedo2022deep}. 
Some diseases could be well represented in such cases, while others have few examples. This can make the model focus on features of more frequent diseases while ignoring the rest.
For example, while a user defines a concept based on his domain knowledge (e.g., disease symptom or texture), the model could instead only focus on a generic feature that is more common across the majority class to make its decision (e.g., Leaf yellowing).

In summary, while significant progress has been made in explainability methods, and their adoption in plant disease classification is increasing, there remains a need for approaches that not only provide visual insights but also uncover high-level concepts and patterns relevant to this domain. This study addresses that need by applying ACE in plant disease classification.

\section{Materials and Methods}
\label{sec:Materials}
\subsection{Plant Disease Classification}
\subsubsection{Model.}
The model selected in this study as the basis for our investigation and experimentation is InceptionV3 \cite{szegedy2015going}. 
This choice is motivated by the wide use of the model in the research community of plant disease classification \cite{fan2022leaf,lee2020new,qiang2019identification,subramanian2022fine}. Also the inception modules use multiple filters that could capture features at different scales which make it suited to capture different patterns present in the plant diseases image dataset. Moreover, the InceptionV3 structure facilitates an easy extraction of the layer and its associated activations. Besides, in this work our primary goal is to understand how these kinds of deep models work rather than focusing on optimizing the classification performance.

To train the network, we used the fine-tuning transfer learning technique \cite{weiss2016survey}.
This is based on transferring the knowledge gained from training the model on a larger data set to a smaller one. Hence, in our case the InceptionV3 model was loaded by the pretrained weights from the ImageNet dataset \cite{deng2009imagenet}. Next new top layers were added to the model. The new layers consisted of a global average pooling layer and three dense layers with a dropout layer. 

 For training and optimizing the weights on the plant disease dataset, we froze the first 52 convolutional layers and made the rest trainable for InceptionV3. Training optimization was carried out via stochastic gradient descent optimizer with a learning rate of 0.0001 and momentum of 0.9. We used a batch size of 64 and 30 epochs for training. 

We use data augmentation techniques to increase the dataset size in the training set while including different variations. This is important since if you train a deep learning model with few examples, it will either underfit by not learning the data well or overfit by not generalising well to not previously seen data. 
In this study, the images are augmented using different transformations such as random rotations, zooms, translations, shears, and flips. 
\subsubsection{PlantVillage dataset.}
The PlantVillage dataset is a public repository that contains 54,323 images of 14 crops and 38 different types of plant diseases \cite{weiss2016survey}. It has been extensively used by the community of plant disease image classification and made a significant contribution to the advancement of applying computer vision and machine learning in agriculture. 

\begin{figure}
  \centering
  \includegraphics[width=\textwidth]{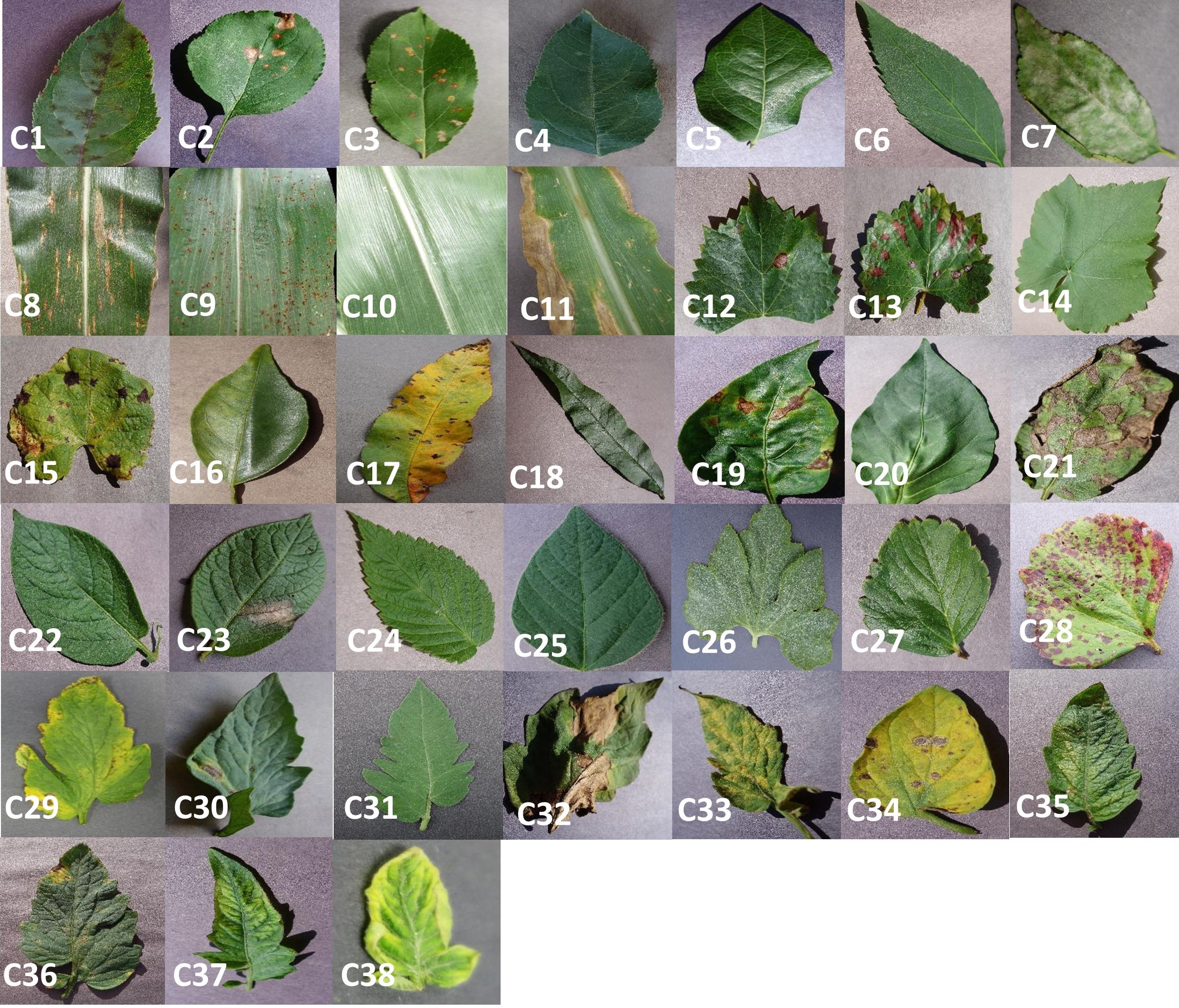}
  \caption{Example of leaf images from the PlantVillage dataset, (C1) Apple Scab, (C2) Apple Black Rot, (C3) Apple Cedar Rust, (C4) Apple Healthy, (C5) Blueberry Healthy, (C6) Cherry Healthy, (C7) Cherry Powdery Mildew, (C8) Corn Gray Leaf Spot, (C9) Corn Common Rust, (C10) Corn Healthy, (C11) Corn Northern Leaf Blight, (C12) Grape Black Rot, (C13) Grape Black Measles(Esca), (C14) Grape Healthy, (C15) Grape Leaf Blight, (C16) Orange Huanglongbing (CitrusGreening), (C17) Peach Bacterial Spot, (C18) Peach Healthy, (C19) Bell Pepper Bacterial Spot, (C20) Bell Pepper Healthy (C21) Potato Early Blight,  (C22) Potato Healthy, (C23) Potato Late Blight,  (C24) Raspberry Healthy (C25) Soybean Healthy, (C26) Squash Powdery Mildew,  (C27) Strawberry Healthy, (C28) Strawberry Leaf Scorch, (C29) Tomato Bacterial Spot,  (C30) Tomato Early Blight, (C31) Tomato Late Blight,  (C32) Tomato Leaf Mold, (C33) Tomato Septoria Leaf Spot,  (C34) Tomato Two Spotted SpiderMite, (C35) Tomato TargetSpot, (C36) Tomato MosaicVirus, (C37) Tomato Yellow LeafCurlVirus, (C38) Tomato Healthy.}
  \label{fig:figplantvillage}
\end{figure}

Figure \ref{fig:figplantvillage} presents an example image of each disease class alongside its corresponding name.
The dataset was initially split into 80\% for training and validation and 20\% for testing. Then, the training and validation portion was further divided, with 80\% used for training and 20\% used for validation.
\subsubsection{Evaluation metrics}
\label{sec:Evaluationmetrics}
To evaluate the performance of the trained InceptionV3 model, we employed several key evaluation metrics: Accuracy, Precision, Recall, and the F1 Score. These metrics together provide a comprehensive view of the model's performance. The full equations of these metrics can be found in the following works \cite{davis2006relationship,hicks2022evaluation}.
\subsection{Network explanation with ACE}
\label{sec:ace}
In their work, Kim et al. \cite{kim2018interpretability} presented concept activation vectors (CAVs) which employ directional derivative and linear separability  to assess the significance of specific user-defined concepts for the deep model functioning.

Based on the foundation laid by \cite{kim2018interpretability}, a subsequent study \cite{ghorbani2019towards} introduced Automatic based concept explanation.  ACE sought to automate the collection of concepts and explore how certain patterns within an image contributed to the model’s decision making mechanism.

\begin{figure}
  \centering
  \includegraphics[width=\textwidth]{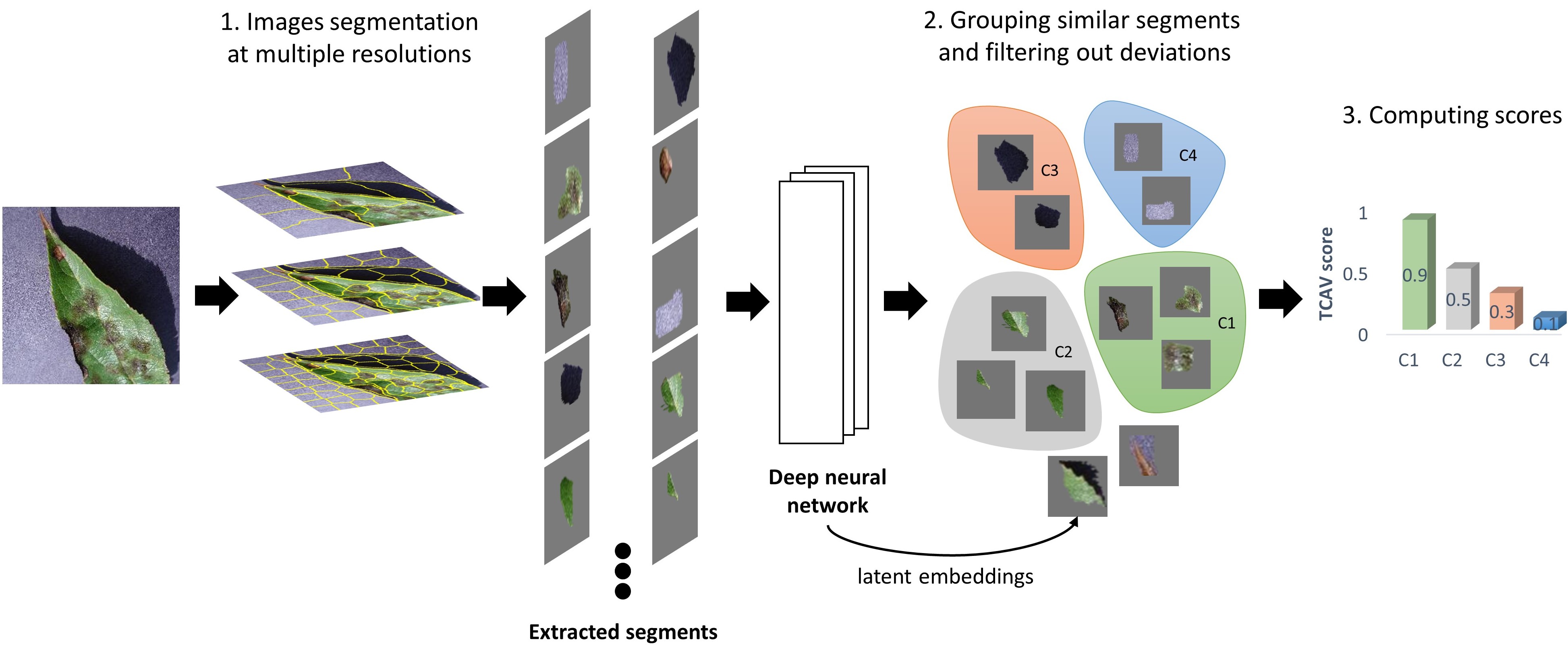}
  \caption{Illustration of the ACE method \cite{ghorbani2019towards} steps for extracting and evaluating plant disease concepts. The first step involves gathering a collection of images belonging to the same plant disease class and applying multi resolution segmentation (SLIC) to each of them. A pool of extracted segments is produced and rescaled to the model’s input size. In the second step, these segments are then passed through the InceptionV3 model, where activations from a  bottleneck layer (latent embeddings) serves as a similarity space. Using these embeddings, k-means clustering groups similar segments into clusters (representing distinct concepts) while removing deviations. In the third step, a TCAV score is computed for each concept to quantify its significance in the model's classification decisions.}
  \label{fig:figframework}
\end{figure}

The ACE method consists of three main steps as shown in Figure \ref{fig:figframework}. The first step consists in conducting a multi resolution segmentation of the images. This means dividing the same image into multiple fragments each time with a different resolution. To achieve this, ACE approach employs the SLIC (Simple Linear Iterative Clustering) algorithm \cite{achanta2012slic}.  SLIC groups pixels that exhibit similar properties such as colour, texture and intensity into clusters called super-pixels. To obtain concepts of varying complexity (coarse to fine), three levels of segmentations are used by ACE (see Figure \ref{fig:figframework}, Step1). 

In the second step of ACE, similar segments are gathered to form examples of the same concept. To achieve this, the intermediate activations of the trained CNN model are used as representation to measure perceptual similarity. This is done by first resizing the segments to the image size adequate and expected by the model and then passing each segment through CNN and measuring the Euclidean distance between their activations in the chosen layer. 

Using k-means clustering, concept patches are gathered into meaningful concept clusters (see Figure \ref{fig:figframework}, Step 2). To ensure concept consistency within each cluster, segments that stand out and  have low similarity to the other segments are removed \cite{ghorbani2019towards}.

Finally, the TCAV scores (see equation \ref{eq2}) for each concept is computed to retrieve the most significant ones for the classification task.
In the following, we will describe the TCAV process in detail. 

Let $k$ be a class label, $X_{k}$ all inputs with this label, $l$ an activation layer from a trained CNN model, $C$ a concept of interest (in our case the concept patches), and $S_{c,k,l}(x)$ the directional derivative. 

First, concept patches are fed to the CNN model up to layer $l$ to extract their activations. Second, these activations are then used to train a linear classifier (SVM) to  differentiate between the concept and random counterexamples. 
The vector orthogonal to the decision boundary separating the two classes, i.e., the vector pointing in the direction of the representations of the concept images, is the CAV ${v_{c}^{l}}$.

 The dot product between the vector ${v_{c}^{l}}$ and the output gradient  $\nabla h_{l,k}$ , which optimises the  prediction of class $k$ is computed to measure the sensitivity $S_{c,k,l}(x)$ (see equation \ref{eq1}) to each concept.
 
 This sensitivity score quantifies how much the classifier's prediction for an input changes when the concept $C$ is present. 
 
The equation representing this sensitivity is provided below.
\begin{equation}\label{eq1}
   \displaystyle  S_{c,k,l}(x) = \lim_{\epsilon  \to 0} \frac{h_{l,k}(f_{l}(x)+\epsilon v_{c}^{l})-h_{l,k}(f_{l}(x))}{\epsilon} \\
    = \nabla h_{l,k}(f_{l}(x)). v_{c}^{l}\\
\end{equation}
Then, to measure the influence of a CAV on a class of input images, a metric called TCAV score is computed. It employs the directional derivatives $S_{c,k,l}(x)$ to compute the contextual sensitivity of a concept towards the whole inputs $X_{k}$ for class $k$. 

Then, to evaluate the impact of a CAV on a specific class of input images, the TCAV score is calculated. This score uses the directional derivatives $S_{c,k,l}(x)$ to determine the overall sensitivity of a concept to the entire set of inputs, $X_{k}$, for class $k$.

The TCAV score is given by:

\begin{equation}\label{eq2}
TCAV_{Q_{c,k,l}}=\frac{\left|x\in X_{k} ; S_{c,k,l}(x)> 0 \right|}{\left| X_{k} \right|}
\end{equation}

Furthermore, to ensure that only meaningful CAVs are considered, the authors perform a statistical significance test on the TCAV scores.

They calculate multiple CAVs by comparing concept images with random images and also train random CAVs, where both the concept and random sets consist of randomly chosen images. 

This ensures that the concept CAVs are statistically distinct from random CAVs, confirming their relevance for class predictions. A two-sided t-test is then applied to the TCAV scores based on these multiple samples. A concept is considered significant for class prediction if the null hypothesis can be rejected. This ensures that the concept CAVs are statistically distinct from random CAVs, confirming their relevance for class predictions.

\subsection{Experimental setup}
\label{sec:Expsetup}
The experiments for both training the model and explainability method were conducted on a server equipped with two NVIDIA Tesla V100 GPUs, each with 16 GB of GPU memory.

The server also has 128 GB of system RAM which ensures enough memory for handling the computational demands of training the deep learning model and the explainability task.
We implemented the model using the Keras \cite{chollet2015keras} deep learning framework.

In this work, we modified ACE method to use it with our trained keras model, incorporating the integration features from the original code. \cite{ghorbani2019towards}.

\section{Results and Discussion}
\label{sec:ResultsandDiscussion}
\subsection{Model training and performance analysis}
In this section, we present the results of training and testing a fine-tuned InceptionV3 model on the PlantVillage dataset. The model was evaluated based on its ability to classify both healthy and diseased plant leaves across multiple species. The performance metrics include accuracy, precision, recall, and F1-scores (see Section \ref{sec:Evaluationmetrics}). After training, our model achieved a training accuracy of 0.98 and  a validation accuracy of 0.94, and a testing accuracy of 0.95. These results demonstrate the model's ability to learn effectively from the training data while maintaining high accuracy on unseen data.

\begin{figure}
  \centering
  \includegraphics[width=\textwidth]{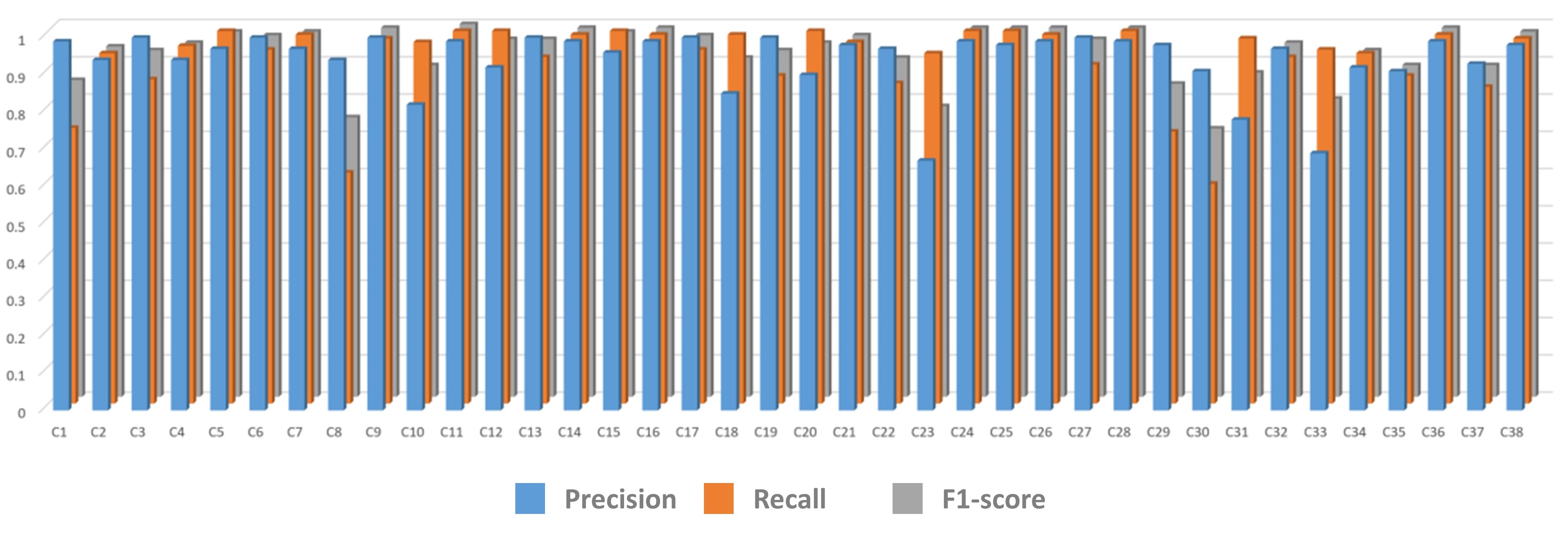}
  \caption{Performance Metrics by Class: Precision, Recall and F1-Score. See Figure \ref{fig:figplantvillage} for the full name of each class.}
  \label{fig:figperformance}
\end{figure}

Figure \ref{fig:figperformance} presents the precision, recall, F1-score for each class, providing a detailed view of the model's performance across each disease.

Some classes exhibit perfect performance with precision, recall, and F1-scores close to or at 1.00 (such as Apple Black Rot (C2), Apple Healthy (C4), and Cherry Powdery mildew (C7)).

However, closer examination reveals that certain classes, particularly those with fewer samples or more complex visual patterns, had notably lower recall and F1-scores. For example, the Apple Scab (C1) class had a recall of 0.74 and an F1-score of 0.85, suggesting that the model struggled to correctly identify all instances of this disease. 

Similarly,  Tomato Early blight (C30) showed a slightly lower performance, with a recall of 0.59 and an F1-score of 0.72.
These differences indicate that while the model performs well on many classes, it struggles to generalise to certain underrepresented or visually complex disease categories.

This observation motivates the need for explainability methods to better understand the model’s decision-making process.  The ACE method is particularly essential here. 
 By analysing the most salient concepts the model uses for classification, ACE can verify whether the model is using expert-relevant concepts such as disease symptoms, color or shape. It also could help in uncovering potential biases or unintended features that the model may rely on for classification.

\subsection{Experiment 1: Insights into the model: Examples of discovered concepts with high and low TCAV scores}

\begin{figure}
  \centering
    \includegraphics[width=\textwidth]{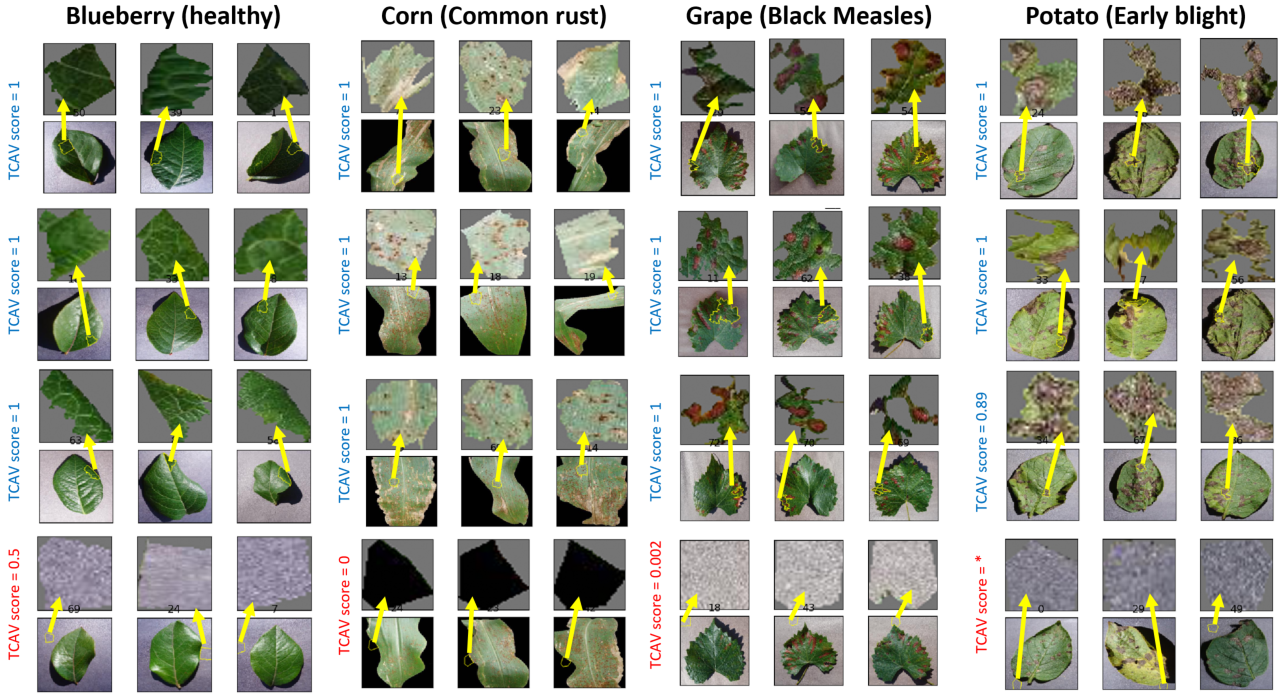}
  \caption{Examples of discovered concepts. For each class, three discovered concepts with High TCAV scores and the one with the lowest score are shown. We randomly chose three segments in each concept.}
  \label{fig:figexperiment1}
\end{figure}

Figure \ref{fig:figexperiment1} presents examples of the concepts discovered by the ACE algorithm, organised from the most to the least salient. 
In each row, the top displays the segmented concepts while the buttom shows the original images from which these segments were extracted.

Each column corresponds to a different class, and within each cell, the discovered concepts  are displayed.
For each class, three discovered concepts with the highest TCAV scores and one with the lowest score are shown. These high TCAV scores indicate that these concepts play a significant role in influencing the neural network’s predictions for that class while it is the opposite for the lowest one. 

To present a comprehensive understanding of these concepts, we have included three randomly selected examples for each concept.
Figure \ref{fig:figexperiment1} shows that ACE has effectively identified key concepts essential to the plant diseases classification, particularly focusing on disease-affected areas of the leaves.

 For each disease class, ACE has captured significant regions that indicate disease symptoms such as spots, blotches and  discoloration deformities.  Notably, a TCAV score of 1 means that 100\% of the images in the target class returned a positive directional derivative, indicating that the discovered concept was highly influential in the network’s predictions \cite{ghorbani2019towards}.
 
On the other hand, for the healthy blueberry class, the identified concepts show different  patches of  leaves with varying vein patterns. This is an interesting observation since  the leaf vein pattern is a critical characteristic to identify the leaf specimen or type by experts \cite{experts2018plant}. The presence of a diverse vein pattern suggests that the model is actively learning the specific variations associated with the leaves in the healthy classes.

This indicates the model's capability to distinguish and differentiate healthy leaf types based on their unique vein configurations, which is a promising feature for accurate plant classification and diagnosis.

\subsection{Experiment 2: Identifying concepts shared across the same disease in different plant species.}
\begin{figure}[h]
  \centering
    \includegraphics[width=\textwidth]{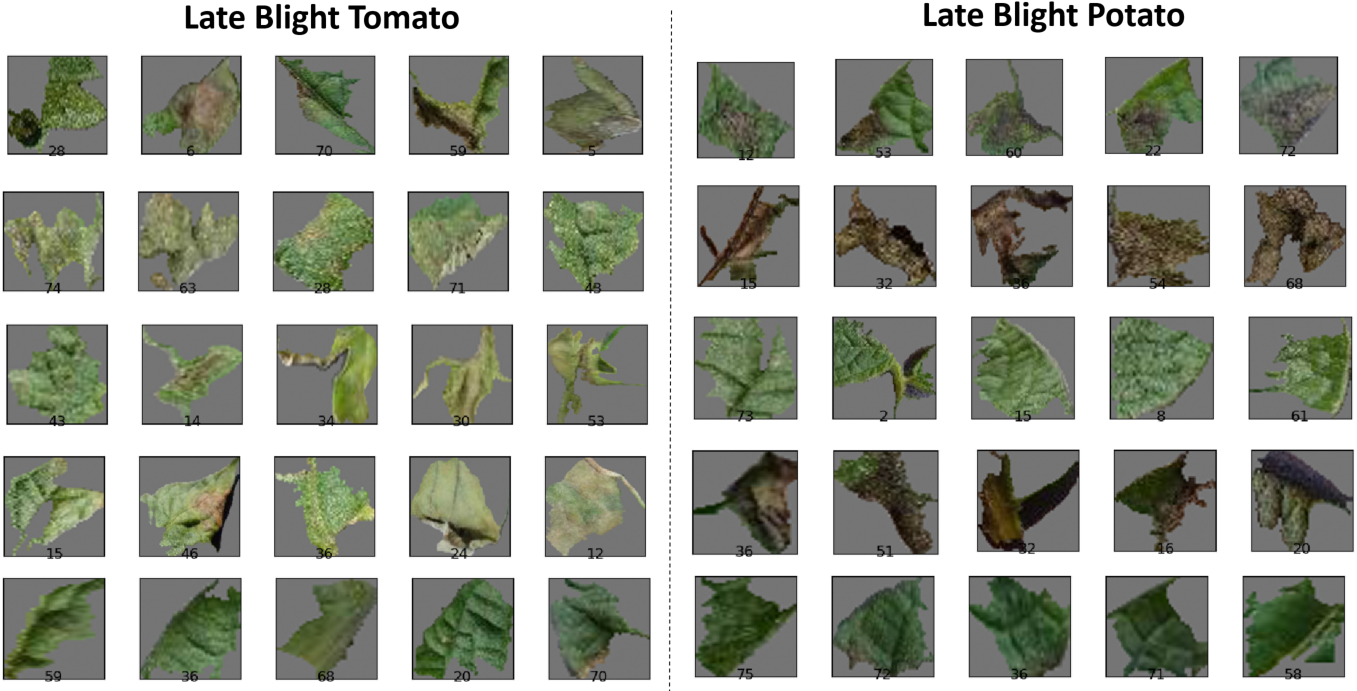}
  \caption{Discovered concepts for the late blight disease in tomato and potato.}
  \label{fig:figexperiment2}
\end{figure}
Another interesting thing is to investigate the concepts the model used when identifying the same type of disease but in different types of plants. This will help us to understand if the model captured the disease pattern regardless of the leaf type.

Late Blight diseases affect both potatoes and tomatoes. Infected leaves often display green to brown areas of dead tissue bordered by a pale green or grey edge. Under conditions of high humidity and wet weather, late blight infections may look water-soaked or dark brown, frequently giving the leaves a greasy appearance \cite{gevens2015tomato}. 

Figure \ref{fig:figexperiment2} shows examples of the learned concepts with high TCAV scores. It seems late blight in Tomato manifest more as pale green patches with some grey hollow and light brown spots. While for Potato, symptoms are more dark brown patches. This could be taken into consideration  when collecting datasets for plant diseases.

This shows how it is important when gathering datasets for plant diseases classification based on images to carefully consider the selection of samples of various plant species affected by the same disease ensuring that the symptoms are alike or close which will help the model to capture the disease pattern regardless of the leaf species.

Another interesting notice is that the model also focused on the venation pattern to be able to also capture the type of the leaf.

\subsection{Experiment 3: Insights into the model from discovered concepts}
Examining the identified concepts that received high TCAV scores provides valuable insights into what the model is focusing on when identifying plant diseases and reveals interesting correlations. Some of these correlations are desirable and in accordance with the expert intuition but others are not and may help in discovering the biases that could be found in the dataset or learned by the model itself.  
\begin{figure}[h]
  \centering
  \includegraphics[width=\textwidth]{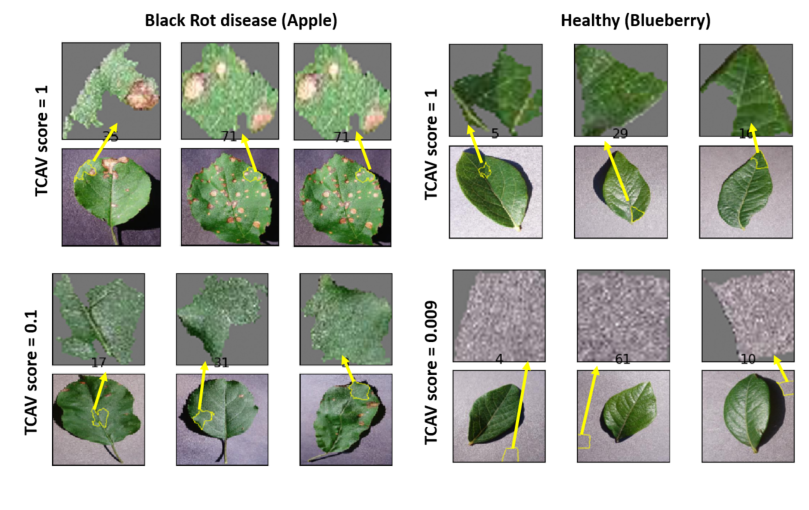}
  \caption{Desirable correlations: distinguishing between Healthy and Diseases leaves}
  \label{fig:figexperiment3_1}
\end{figure}
In the following sections, these findings will be studied in detail.
\subsubsection{A. Desirable correlations: distinguishing between Healthy and Diseases leaves}
Figure \ref{fig:figexperiment3_1} reveals some of the desirable correlations found. In the Black rot disease class, regions of the leaf with disease spots had significantly higher TCAV score than the unaffected parts of the leaf. This means that the disease spots had a big impact on the model decision making process while the unaffected parts had less of an impact. This shows the model’s ability to effectively differentiate between healthy and disease patterns and make informed classification.

In the healthy blueberry class, we see that the model captured contextually relevant concepts. For instance, the healthy patches of the leaves were identified as significant concepts while the background of the leaf was correctly recognized as not important for the classification task. 

This could be used as a demonstration for the network’s ability in learning the important features associated with the corresponding class and ignoring the rest.

\subsubsection{B. Desirable correlations: healthy classes and venation patterns}
\begin{figure}
  \centering
  \includegraphics[width=0.7\textwidth]{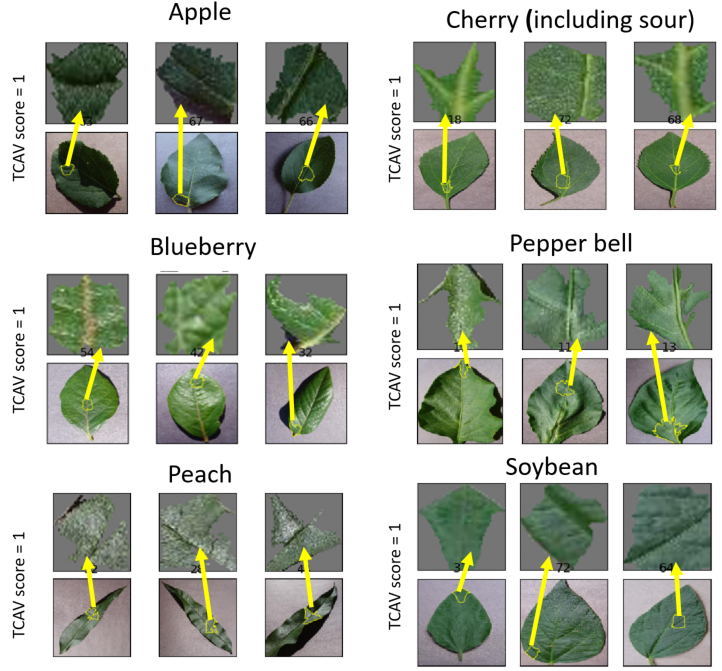}
  \caption{Desirable correlations: healthy classes and venation patterns.}
  \label{fig:figexperiment3_2}
\end{figure}
Another interesting insight was found when checking most significant extracted concepts (TCAV score: 1.0) for healthy classes of different plants within the dataset (see Figure \ref{fig:figexperiment3_2}).

The identified concepts show different  patches of leaves with varying vein patterns. This is an interesting observation since  the leaf vein pattern is a critical characteristic to identify the leaf specimen or type by experts. The presence of a diverse vein pattern suggests that the model is actively learning the specific variations associated with the leaves in the healthy classes.

This indicates the model's capability to distinguish and differentiate healthy leaf types based on their unique vein configurations, which is a promising feature for accurate plant classification and diagnosis.
This shows that the identified concepts  are aligned with expert intuition in identifying leaves types.

\subsubsection{C. Undesirable correlations: Bias detection}
ACE helped to discover some of the undesirable correlations learned by the model and identify the bias within the dataset it was trained on notably concerning background and shadow factors.
\begin{itemize}
 \item \textbf{Background Bias:} 
  \begin{figure}
  \centering
  \includegraphics[width=0.7\textwidth]{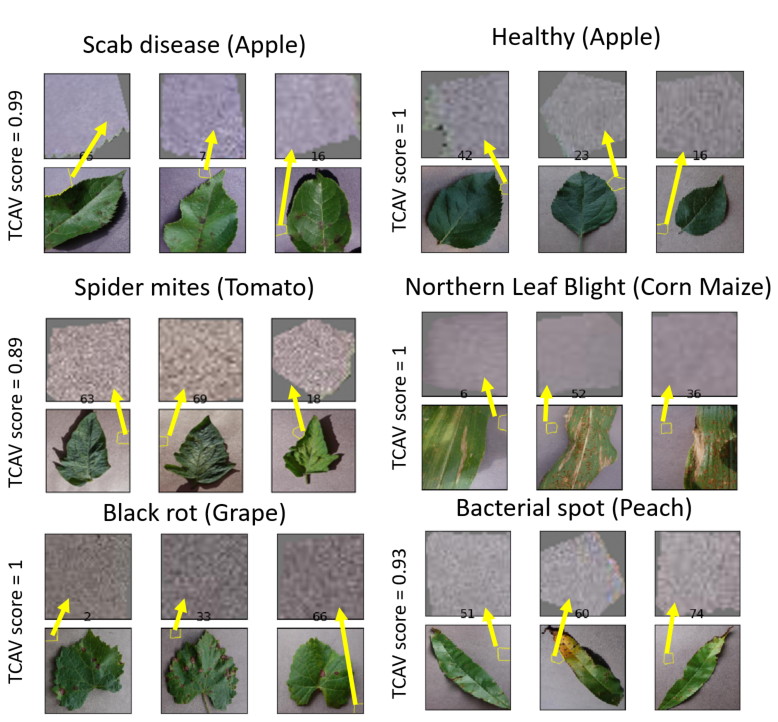}
  \caption{Background bias.}
  \label{fig:figbackgroundbias}
\end{figure}
Figure \ref{fig:figbackgroundbias} highlights the presence of background bias. This is a phenomenon where the model mistakenly associates certain background colors or patterns with specific classes. 

Instead of focusing on the main object which is in our case the leaf type and the disease patterns, the model uses the background features as a distinguishing component for classifying the corresponding class. 

This bias occurs when images of a particular class frequently appear against a specific type of backgrounds. Thus, the model might begin to correlate the background color or texture with that class. ACE has successfully detected this issue by identifying variations in background colors across different images, which may have influenced the model’s decision-making process.

For example, for the Scab Disease (Apple) the background apears to be light purple and uniform across different images while for Spider mites (Tomato), the background appears darker and grey-brown. These differences in color and brightness if the background across the different classes may be leading the model to unintentionally using it as indicators of the classes.
 This finding is significant, as it highlights how unintended elements of an image, like the background, can create misleading associations.

 \item \textbf{Shadow Bias/ Light Bias:}
  \begin{figure}
  \centering
  \includegraphics[width=0.7\textwidth]{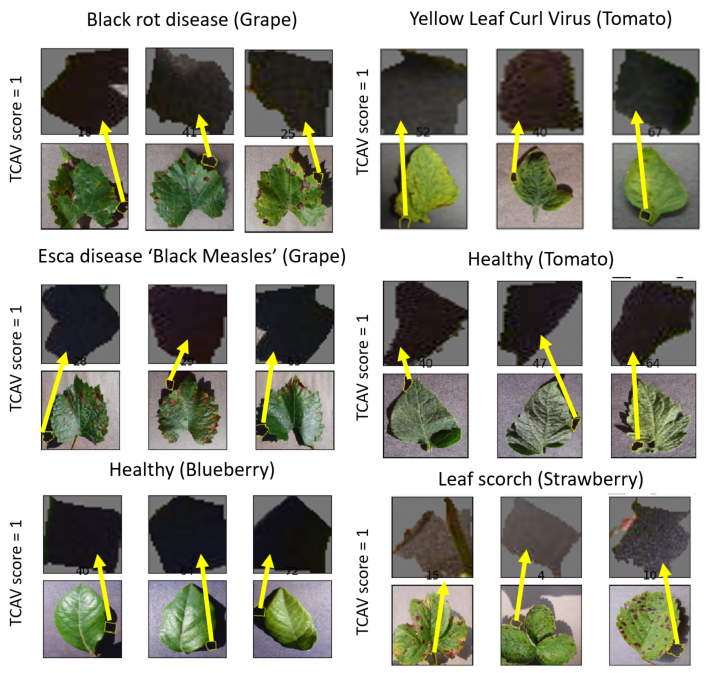}
  \caption{Shadow Bias.}
  \label{fig:figexperiment3_4}
\end{figure}

 Another bias that was identified through the discovered concepts is the shadow of the leaves on the background, which occurred when the leaves were photographed under certain lighting conditions. This could lead the model to  incorrectly associating the dark shadows with disease symptoms, particularly those that are marked by black or dark-coloured spots.
For example, diseases like black rot (see Figure \ref{fig:figexperiment3_4}) or other infections that cause dark patches on leaves could become confused with the black shadows in the background.
Also this bias seems to not just affect the diseased classes but also the healthy classes (see Figure \ref{fig:figexperiment3_4} Tomato and Blueberry healthy examples).  The model may have learned to associate the dark shadows in the background as a feature relevant to both diseased and healthy classes, leading to misclassification. 

This highlights the value of using the ACE method for model inspection. By identifying such biases, researchers can take corrective actions to improve the dataset’s representation. These actions may include augmenting the dataset with more diverse samples and refining image capture techniques to eliminate unintended influences like shadows or background variations.

 Such improvements help ensure that the model focuses on the biological relevant traits and concepts for plant diseases classification which will lead to more accurate and reliable classifications.

\end{itemize}
\subsection{Experiment 4: Lowest recall and F1 score classes discovered concepts}

Another interesting insight appeared when we checked the most salient concepts for classes with low recall and F1-scores. We found that these concepts were mostly related to the background rather than the actual disease features. This suggests that the model became confused and failed to learn the correct patterns associated with these classes, instead focusing more on the background as a distinguishing feature. As a result, this contributed to the poor performance in accurately detecting them.

This finding indicates that the classifier may struggle with these specific classes since it doesn't effectively identify the relevant disease features. To address this, incorporating additional training data with more varied and diverse backgrounds could help the model learn to focus on the leaf characteristics rather than the background. Additionally, checking for class imbalance could be beneficial, as imbalanced data may be further affecting the model's ability to correctly classify these classes. For instance, due to dataset constraints, only 31 images were available for testing in the Potao Healthy class. Ensuring balanced representation across all classes, combined with background diversity, could significantly improve the model’s performance.
\begin{figure}
  \centering
  \includegraphics[width=\textwidth]{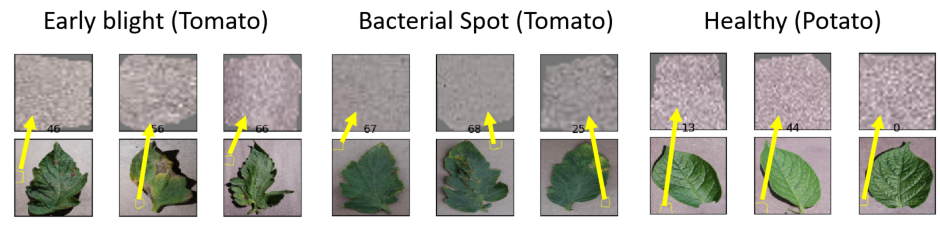}
  \caption{Examples of discovered concepts in classes with the lowest recall and F1-scores, showing high TCAV scores.}
  \label{fig:figexperiment4}
\end{figure}

\section{Conclusion}

In this work, we present the first study of Automated Concept-based Explanation (ACE) to deep learning-based plant disease classification to enhance the explainability of deep learning in agricultural disease diagnosis. 
Our objective is to uncover the specific visual concepts that the model uses for its decisions. 
This is an essential step toward developing transparent and trustworthy tools for plant disease management. Using the well-known InceptionV3 model and PlantVillage dataset, we demonstrate how ACE can automatically identify significant visual patterns that influence the model’s decision-making.

Through a series of experiments, we demonstrated ACE’s ability to provide valuable insights into plant disease classification by automatically identifying key visual concepts, such as spots or infected areas,  while also revealing potential sources of error, including misleading correlations with background or shadow elements.
 Furthermore, ACE also highlighted areas where the model struggled, particularly in classes with lower recall and F1-scores, which allows us to identify where targeted improvements could be most beneficial.
Hence, this approach holds significant value for various stakeholders. It will increase users' trust, such as plant experts, agriculturists, and farmers, by clarifying why predictions are made. It also supports data scientists and deep learning researchers in diagnosing issues and developing more robust models.
In future work, we aim to apply ACE to larger and more diverse datasets that include real-world backgrounds. Also, we would like to test its applicability to other deep learning architectures. Furthermore, integrating ACE into a real-time tool could enable users to explore and validate concept clusters interactively. Such an approach would increase the practicality of ACE and thus support users in refining and improving the model's accuracy and trustworthiness.

\begin{credits}
\subsubsection{\ackname}  This work was supported by the Carl Zeiss Foundation (project ‘A Virtual Werkstatt for Digitization in the Sciences (K3)’ within the scope of the programline ‘Breakthroughs: Exploring Intelligent Systems for Digitization - explore the basics, use applications’). The computational experiments were performed on resources of Friedrich Schiller University Jena supported in part by DFG grants INST 275/334-1 FUGG and INST 275/363-1 FUGG. During the preparation of this work, the first author used ChatGPT to check certain sentences for grammar and clarity. After using this service, all authors reviewed and edited the content as needed and took full responsibility for the publication's content.
\subsubsection{CRediT authorship contribution statement.} \textbf{Jihen Amara}: Conceptualization, Methodology, Investigation, Writing – original draft. \textbf{Birgitta König-Ries}: Validation, Supervision, Writing - Review \& Editing. \textbf{Sheeba Samuel}: Writing - Review \& Editing.

\end{credits}
%
%
%
 \bibliographystyle{splncs04}
 \bibliography{mybibliography}

\end{document}